\documentclass{article}
\newif\ifpreprint
\preprinttrue
\usepackage{iclr2027_conference,times}

\usepackage{url}
\usepackage[T1]{fontenc}
\usepackage{amsmath,amssymb}
\usepackage{booktabs}
\usepackage{graphicx}
\usepackage{float}
\usepackage{placeins}
\usepackage{tabularx}
\usepackage{adjustbox}
\usepackage{xcolor}
\usepackage[hidelinks]{hyperref}
\usepackage{dblfloatfix}

\newcommand{\raw}{\mathrm{raw}}
\newcommand{\comp}{\mathrm{comp}}
\newcommand{\pp}{pp}
\newcommand{\tablefont}{\fontsize{8}{9.2}\selectfont}
\definecolor{benchhotpot}{HTML}{2F6F9F}
\definecolor{benchmusique}{HTML}{A65F00}
\definecolor{benchother}{HTML}{8A8A8A}
\newcommand{\benchswatch}[1]{\textcolor{#1}{\rule[-0.1ex]{1.1ex}{1.1ex}}\hspace{0.45em}}
\newcommand{\retbar}[2]{\rlap{\textcolor{#1!22}{\rule[-0.35ex]{#2\dimexpr0.015cm\relax}{1.75ex}}}\makebox[1.5cm][r]{#2}}

\title{Compression Is Not Evaluation-Neutral:\\
Fixed RAG Compression Can Distort Reader Comparisons}

\author{Anonymous authors}

\ifpreprint
  \iclrfinalcopy
  \usepackage{aimsheader}

  \hypersetup{
    pdftitle={Compression Is Not Evaluation-Neutral: Fixed RAG Compression Can Distort Reader Comparisons},
    pdfauthor={Sugam Panthi, Rabab Abdelfattah}
  }
  \newcommand{\codeurl}{https://github.com/aimsresearchlab/ragscale}
  \fancypagestyle{plain}{\fancyhf{}\cfoot{\thepage}\renewcommand{\headrulewidth}{0pt}}
\fi

\begin{document}
\ifpreprint
  \aimslinetitle{
    title   = {\fontsize{18}{22}\selectfont Compression Is Not Evaluation-Neutral:\\
      Fixed RAG Compression Can Distort Reader Comparisons},
    authors = {Sugam Panthi, Rabab Abdelfattah},
    affil   = {AIMS Lab, The University of Southern Mississippi\\
      \texttt{\{sugam.panthi, rabab.abdelfattah\}@usm.edu}},
    right   = {Preprint \textbullet\ October 2026},
  }
  \thispagestyle{plain}
\fi

\begin{abstract}
Retrieval-augmented generation (RAG) evaluations often compare readers after a
compressor has changed their evidence. This mixes two questions: which complete
pipeline works best, and how much of a reader upgrade survives
compression. We show that fixed compression can raise average pipeline accuracy
while hiding most of a reader upgrade. We hold questions, retrieved candidates,
prompts, scoring, and compressed text fixed while comparing 8--20 readers
across five question-answering benchmarks and five compression families.
HotpotQA and MuSiQue are the main benchmarks, analyzed under a plan fixed in
advance. In the 20-reader HotpotQA
panel, the lowest- and highest-scoring readers under raw evidence are 31.8
percentage points (pp) apart before compression but only 7.8\pp{} apart on the
same stored output from a HotpotQA-trained RECOMP compressor. On both main
benchmarks, lower raw-scoring readers gain more under exact match
and token-overlap F1. Reader pairs also change order more often under compressed
evidence than in raw replays across question halves. Row-level accounting
explains how higher average accuracy and
smaller upgrades coexist: compression rescues some wrong answers and
damages some correct answers, with a different balance for each reader. We
release \texttt{ragscale} to audit declared reader upgrades under raw and fixed
compressed evidence. Reader evaluations should run this audit before
attributing a compressed-only result to the reader.
\end{abstract}

\ifpreprint
\begin{center}
\small\mbox{\textbf{Code and data:} \texttt{ragscale} and the interaction matrix at
\href{\codeurl}{\raisebox{-0.2em}{\includegraphics[height=1.1em]{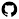}}\,\texttt{aimsresearchlab/ragscale}}}
\end{center}
\fi

\section{Introduction}
\label{sec:introduction}

Retrieval-augmented generation (RAG) retrieves passages before an answer model
answers a question. We call the answer model the reader, and a system that
shortens or rewrites the passages a compressor. An evaluation can ask which
complete pipeline works best, or whether a reader
upgrade survives the compression used to measure it. These are different
evaluation targets. Compression is usually judged by whether the shorter
pipeline preserves or improves answer accuracy
\citep{xu2024recomp,jiang2024longllmlingua,pan2024llmlingua2,pisco2025}. But the
same compressor can help the full system while hiding a reader upgrade. On
HotpotQA, the raw upgrade from the lowest- to the highest-scoring reader is
31.8\pp{} but falls to 7.8\pp{} on one stored set of RECOMP outputs
\citep{xu2024recomp,yang2018hotpotqa}. RECOMP is a compressor trained on
HotpotQA. It raises several pipeline scores while three quarters of the
upgrade disappears.

\begin{figure*}[t]
  \centering
  \includegraphics[width=0.94\textwidth]{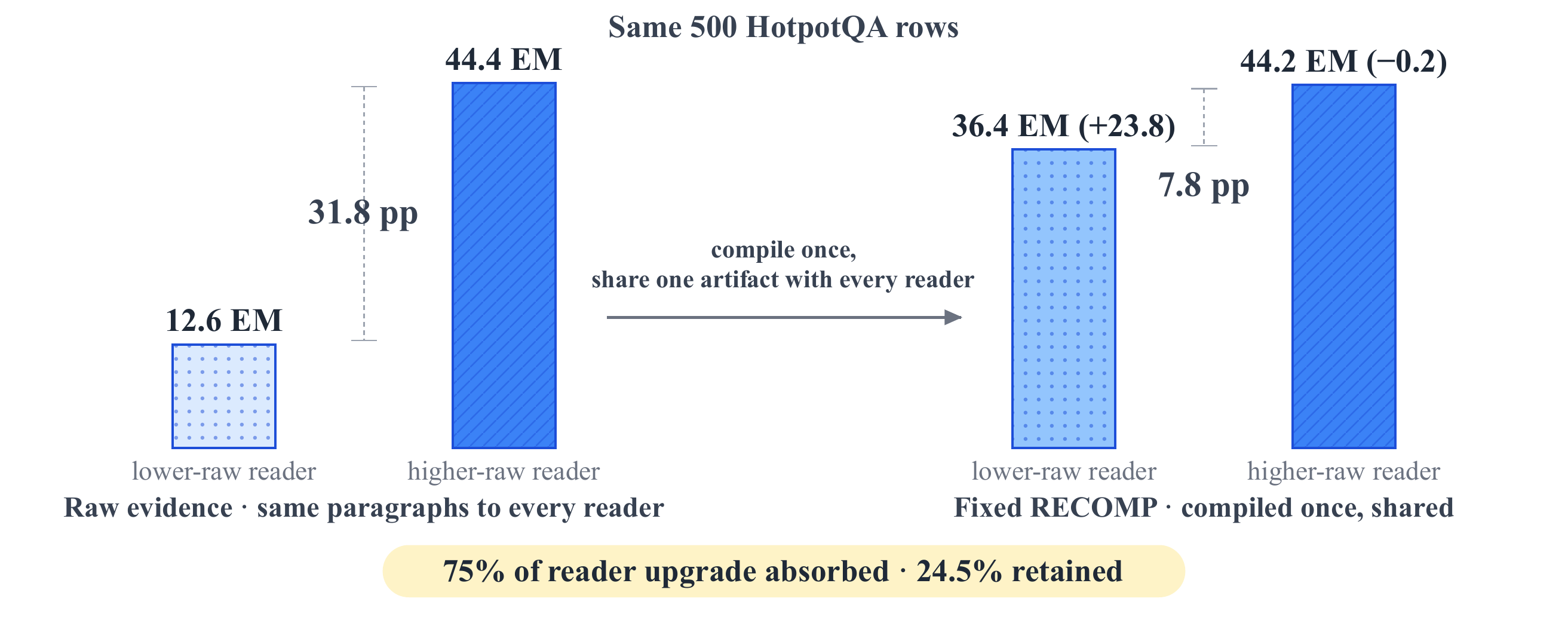}
  \caption{\textbf{One useful compressor changes the reader comparison.}
  The same 500 HotpotQA candidate pools produce raw evidence and one stored
  RECOMP version. Every reader receives exactly the same compressed text. For
  the readers with the lowest and highest raw scores, RECOMP raises the lower
  score by 23.8\pp{} but changes the higher score by $-0.2$\pp{}. Their visible
  reader upgrade falls from 31.8\pp{} to 7.8\pp{}, so 75\% disappears. RECOMP
  helps the lower-scoring pipeline without preserving the reader comparison.}
  \label{fig:reader-replay}
\end{figure*}

If a deployed system always uses one compressor, its compressed score is the
right deployment target. However, that score alone does not show whether the compressor
preserved the reader comparison. Compression can raise the deployed score while
making an upgrade look smaller or a different reader appear better. When a team replaces the reader but leaves
retrieval and compression unchanged, it therefore needs both raw and compressed
results. A compressed-only A/B test can hide an upgrade that is clear under raw
evidence.

Prior work establishes that compression can shorten context while preserving
or improving answer accuracy. It also reports mixed effects across readers
\citep{exit2025,refiner2024}. However, its cross-reader displays contain only
2--4 readers and do not verify that every reader received identical compressed
text (Section~\ref{sec:related}; Appendix~\ref{app:prior-audit}). A shared method
name is not enough: regenerating a summary can give two readers different
evidence. The remaining question is therefore whether one fixed compression output
preserves a broad reader comparison.

We make this comparison direct. Within each panel, we keep
the questions, retrieved candidate passages, prompt, readers, and scoring rule
unchanged. We store one compressed output for each question and give that exact
text to every reader. Each reader answers with the raw candidate passages and
with the same stored compressed version. \emph{Upgrade retention} measures how
much of the raw upgrade remains after compression. For example, a
10\pp{} raw upgrade that falls to 3\pp{} has 30\% retention. We also ask
whether compression makes separated reader comparisons less stable. As a secondary diagnostic, we
compare observed sign changes with those produced by different random halves
of the raw examples. These random halves estimate how often row sampling alone
changes an ordering.

HotpotQA and MuSiQue \citep{trivedi2022musique} both show attenuation under
their standard answer metrics.
HotpotQA retains 24.5\% of its raw upgrade under RECOMP. On MuSiQue, lower
raw-scoring readers gain more under both token-overlap F1 and exact match, and
four stored summaries each retain 54--76\% of the upgrade across EM and F1 on a shared
ten-reader panel. Smaller upgrades are also harder to resolve: only 41.7\% of
HotpotQA reader pairs that differ significantly under raw evidence still 
differ in the same direction under compression. Finally, a row-by-row accounting shows how higher
average accuracy and smaller upgrades coexist. Compression
rescues some previously wrong answers and damages some previously correct
answers, and the balance differs across readers.

The full study spans five question-answering benchmarks, five compression
families, and panels of 8--20 readers. HotpotQA, MuSiQue, LongMemEval, and
NQ-Open form the primary study. A later TriviaQA panel adds a robustness check
\citep{joshi-etal-2017-triviaqa}. The
effect varies by compressor: on the same HotpotQA questions and readers, EXIT
retains 83.6\% of the upgrade under exact match.

We make three contributions.
\begin{enumerate}
  \item We separate two questions that compressed evaluations mix: which
  pipeline works best, and how much of a reader upgrade survives compression.
  We measure the second by giving every reader the same stored compressed text.
  \item We show that fixed compression shrinks reader upgrades on HotpotQA and
  MuSiQue, and that reader pairs become harder to tell apart.
  \item We release a dataset of reader answers under raw and compressed
  evidence, and \texttt{ragscale}, a pip-installable toolkit that runs this
  comparison for future systems.
\end{enumerate}

Reader evaluations should give every reader the same compressed text, report raw and
compressed reader results, and show how much of each upgrade
survives.

\section{Related Work}
\label{sec:related}

\paragraph{RAG evidence compression.}
Extractive compressors filter sentences or passages
\citep{wang2023filco,xu2024recomp,zhao2024seer}. Other methods synthesize
retrieved documents, prune tokens, or encode evidence into latent
representations \citep{compact2024,pisco2025,jiang2024longllmlingua,
pan2024llmlingua2,cheng2024xrag,ge2024icae}.
Most work asks whether the resulting pipeline stays accurate while using less
context \citep{li2024promptcompressionsurvey,jha2024characterizing}. By contrast, we ask: when the compressed evidence is held
fixed, does it preserve the reader upgrade?

Production-oriented and research RAG tooling makes evidence preparation a
reusable stage \citep{rau2024bergen,jin2025flashrag}. LangChain wraps a
retriever with a document compressor, GraphRAG answers from generated community
reports, and AWS describes query-aware compression to lower input cost
\citep{langchain2026compression,microsoft2026graphrag,aws2026ragcompression}.
This stage can stay fixed when a team changes its reader. In that case, a
compressed pipeline score alone cannot show how much of that upgrade remains visible.

\paragraph{Comparisons across readers.}
EXIT and Refiner provide the closest prior observations. EXIT reports larger
gains for a 70B reader, while Refiner reports larger gains for a weaker reader
and degradation for a stronger one \citep{exit2025,refiner2024}. However, our audit of
eight compression papers finds that their primary cross-reader displays contain
2--4 readers, with a median of three. None reports enough provenance to verify
identical artifacts across readers. These mixed results motivate a controlled
study, but they cannot establish whether one fixed artifact preserves reader
comparisons (Appendix~\ref{app:prior-audit}).

Related work also varies compressor size and measures reconstruction fidelity
\citep{sizefidelity2026}. Our experiments hold the compressed artifact fixed
and vary the answer reader.

\paragraph{Readers under changing evidence.}
Context length, evidence position, irrelevant passages, and hard negatives can
all change reader accuracy
\citep{levy2024context,yoran2024irrelevant,jin2025longcontext,
liu2024lostinthemiddle}.
Mechanistic work also shows that readers use different internal processes to
retrieve information from context \citep{wu2025retrievalhead}. These findings
explain why compression can affect readers differently. However, they do not measure
whether a shared compressor preserves the comparison between them.

\paragraph{Compression as an evaluation intervention.}
Compression research measures several objectives, including retained
information, latency, rate adherence, and downstream accuracy. These objectives
need not agree, and pruning can keep a relevant fragment while dropping context
needed to interpret it
\citep{nagle2024fundamental,arda2025ratedistortionsummarization,
lajewska2025informationpreservation,kummer2026compressionwild,
johnson2026compressionmethod,hu2026referentialdangling}. Evaluation studies
likewise show that formatting or scoring choices can change model scores and
rankings \citep{sclar2024formatspread,mirzadeh2025gsmsymbolic,
su2025singlecharacter}. Closest to our framing,
\citet{panthi2026sameranking} show that changing the scoring target in memory
benchmarks can reverse which system wins. Compression is therefore both a
deployment component and an intervention on the comparison being measured.

\section{Measuring Reader Comparisons Under Compression}
\label{sec:estimands}

Compression can change a reader's own score and the visible upgrade from one
reader to another. These are different outcomes. The first describes the
compressed pipeline. The second tells us whether compression preserves the
reader comparison. We define them separately.

For each question, we store the full candidate pool and a reference answer. The
benchmark setting determines which passages form the raw reader input. The
compressor turns the stored pool into one compressed output, and every reader
receives that exact text. Raw and compressed conditions therefore share the
question, candidate pool, prompt, readers, and scoring. They differ only in the
evidence passed to the reader.

Formally, row $i$ contains question $q_i$, stored candidate pool $z_i$, raw
reader input $x_i$, and reference answer $y_i$. The benchmark setting takes
$x_i$ from $z_i$, while the compressor produces
$\widetilde{x}_i=c(q_i,z_i)$. We omit the shared prompt from the notation. The
raw input applies no learned compression or summarization, although its
retrieval cutoff can differ by benchmark. It defines our comparison baseline.

For score function $s$, reader $r$ has raw and compressed means
\begin{align}
  \mu_r^{\raw} &= \frac{1}{n}\sum_{i=1}^{n}s(r,q_i,x_i,y_i), &
  \mu_r^{\comp} &= \frac{1}{n}\sum_{i=1}^{n}s(r,q_i,\widetilde{x}_i,y_i).
\end{align}
The difference $\mu_r^{\comp}-\mu_r^{\raw}$ is the reader's compression gain.
A positive value means that the compressed pipeline helps reader $r$. Whether
a reader comparison survives depends on how both readers' scores change.

For a declared change from reader $a$ to reader $b$, the visible reader upgrade
under evidence condition $p$ is
\begin{equation}
  \Delta^p_{a\rightarrow b}=\mu_b^p-\mu_a^p,
  \qquad p\in\{\raw,\comp\}.
\end{equation}
Compression preserves the upgrade exactly when both readers receive the same
gain. If reader $a$ gains more than reader $b$, the upgrade shrinks and can
eventually reverse.

For readers ordered so that $b$ has the higher raw score, \emph{upgrade
retention} is
\begin{equation}
  \rho_{a\rightarrow b}=
  \frac{\Delta^{\comp}_{a\rightarrow b}}
       {\Delta^{\raw}_{a\rightarrow b}}.
\end{equation}
A value of 1 preserves the raw upgrade. A value between 0 and 1 shrinks it, 0
removes it, a negative value reverses it, and a value above 1 enlarges it.

For the headline result, we choose the readers with the lowest and highest raw
scores. We keep that pair fixed under compression. Because retention is unstable when the raw upgrade is
small, resampled pairs must remain at least 5\pp{} apart under raw evidence. As
a sensitivity check, we select the pair on one random half of the questions
and score it on the other.

Close readers can change order simply because a benchmark contains a finite
set of questions. We estimate this background rate by repeatedly dividing the
questions into two random halves. On the first half, we select every reader
pair separated by at least 5\pp{} under raw evidence and record which reader
leads. On the second half, we ask how often those orderings reverse under raw
and compressed evidence. We repeat this comparison across 5,000 splits.

We call the compressed reversal rate minus the raw reversal rate \emph{excess
flips}. A positive value means that reader orderings reverse more often under
compression than they do after changing only the sampled raw questions.
Upgrade retention measures how much of a raw upgrade remains. Excess
flips, in contrast, measure additional observed reversals beyond the raw row-sampling floor.
Upgrade attenuation alone can produce excess flips, so they do not by themselves
establish a different underlying reader ordering.

\section{Experimental Design}
\label{sec:design}

Each panel stores one compressor output per question and gives that same text to
every reader. Content hashes link each output to its candidate pool. We then
apply the measures in Section~\ref{sec:estimands} to readers with complete raw
and compressed results on those questions.

\paragraph{One exact artifact per panel.}
A compression method enters a panel only when every reader receives the same
stored artifact on every row. Content hashes verify this identity. We intersect
raw and compressed row identifiers before scoring and never impute a missing
score. Generated summaries can change when created again for a later reader
batch. The stored text therefore defines the experimental control.

\paragraph{Benchmarks and candidate pools.}
LongMemEval retrieves a BM25 top-20 pool from roughly 48 conversation sessions.
The raw reader receives that pool directly, while each compressor transforms
the same pool. HotpotQA uses all ten paragraphs from the distractor setting in
their supplied order. RECOMP first ranks those paragraphs with BM25, then its
in-domain checkpoint compresses the top five into the fixed artifact. Ties keep
the supplied order.

MuSiQue supplies up to twenty paragraphs, about 90\% of which are
distractors. Its canonical raw input contains the first eight paragraphs in
supplied order, while the summary compiler can use all twenty. Both inputs come
from the same stored pool, but the summary can draw from more of it. A
matched-input check therefore gives three raw readers every supplied paragraph.
This check isolates unequal candidate access for those three readers. Length
and format still differ between the raw passages and the summaries.

NQ-Open uses one stored open-domain pool per question. TriviaQA retrieves a
BM25 top-20 pool from its supplied Wikipedia pages. Across the study, the five
compressor families are RECOMP, EXIT, Provence, SIEVE, and shared generated
summaries. The five NQ-Open settings use RECOMP top-5, RECOMP with raw fallback,
EXIT, Provence, and one shared summary.

\paragraph{Scope and evidential roles.}
Table~\ref{tab:scope} states what each benchmark contributes. The two
confirmatory rows use native answer metrics. The remaining rows test semantic
replication, a lexical boundary, or later robustness.

\begin{table}[H]
\centering
\tablefont
\setlength{\tabcolsep}{3.5pt}
\begin{tabularx}{\textwidth}{@{}lrrXXll@{}}
\toprule
Benchmark & Rows & Readers & Raw reader input & Fixed compressor & Metric & Role \\
\midrule
\benchswatch{benchhotpot}HotpotQA & 500 & 20 & 10 supplied paragraphs & in-domain RECOMP top-5 & EM/F1 & Confirmatory \\
\benchswatch{benchmusique}MuSiQue & 500 & 15 & first 8 of up to 20 paragraphs & one summary from up to 20 & EM/F1 & Confirmatory \\
\benchswatch{benchother}LongMemEval & 500 & 17--20 & BM25 top-20 & SIEVE or shared summary & Semantic & Replication \\
\benchswatch{benchother}NQ-Open & 323 & 8--20 & one fixed open-domain pool & five fixed settings & Lexical & Boundary \\
\benchswatch{benchother}TriviaQA$^{\dagger}$ & 500 & 13 & BM25 top-20 from supplied Wikipedia & one shared summary & EM/F1 & Robustness \\
\bottomrule
\end{tabularx}
\caption{\textbf{Five benchmarks play different roles in the evidence.}
$\dagger$TriviaQA was added as a later robustness check after the HotpotQA and MuSiQue
analysis plan was fixed.}
\label{tab:scope}
\end{table}

\paragraph{Readers and released artifact.}
The panels contain 8--20 readers from as many as eleven model families. They
range from 7B open models to hosted frontier systems.

We release 176,864 reader-by-compression rows with the candidate pools,
compressed evidence, reader generations, scores, content hashes, and audit
diagnostics. The source benchmarks remain separately cited datasets. The later
TriviaQA runs sit outside this primary matrix.

\paragraph{Scores and uncertainty.}
HotpotQA and MuSiQue use benchmark-compatible exact match (EM) and token F1.
TriviaQA takes the best EM and F1 over its supplied answer aliases. LongMemEval
uses its fixed semantic-judge rubric. NQ-Open also takes the best normalized
match over its accepted aliases. This lexical score differs from Google's
document-level NQ evaluator.

All interval analyses use 5,000 resamples. Family-by-row intervals resample
questions and reader families together. They describe uncertainty over the
sampled questions and represented reader families. Ordering intervals show
sensitivity across disjoint row splits of the fixed panel.

Raw score also appears inside compression gain, so their observed correlation
shares row-sampling noise. We estimate and remove that shared noise before
reporting the correlation. Appendix~\ref{app:replication} gives the formula,
scoring rules, reader inventory, and sensitivity settings. Methods within one
benchmark share questions and readers. We treat those methods as related checks
within one benchmark-level replication.

\FloatBarrier

\section{Fixed Compression Attenuates Reader Upgrades}
\label{sec:effects}

\subsection{Both confirmatory benchmarks show attenuation}

\begin{figure}[H]
  \centering
  \includegraphics[width=0.84\textwidth]{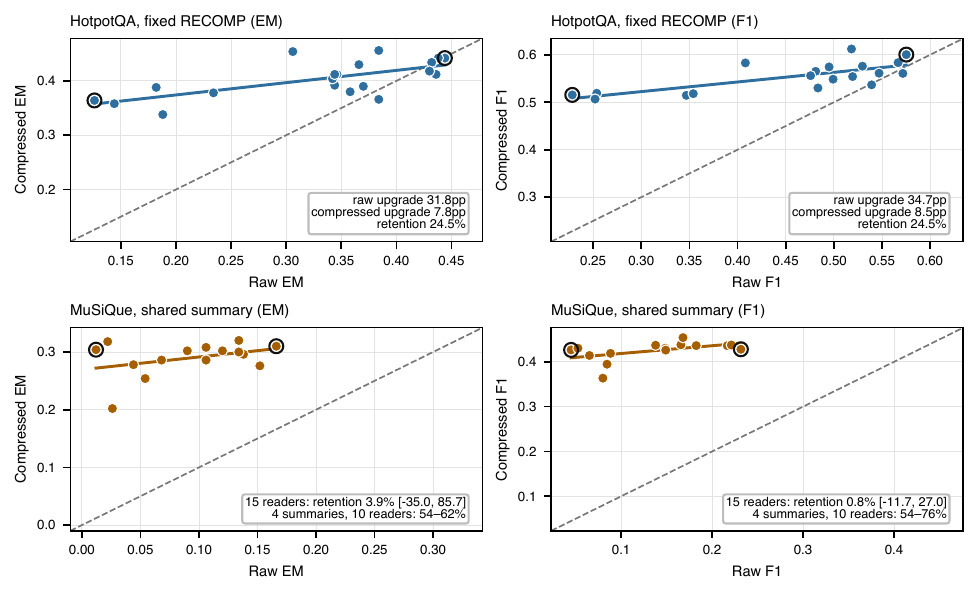}
  \caption{\textbf{Fixed compression flattens the visible reader-scaling
  curve.} Each point is one reader under raw and byte-identical compressed
  evidence. Solid lines summarize each panel; the dashed diagonal marks equal
  raw and compressed scores. Rings mark the lowest- and highest-scoring raw readers. For MuSiQue, the
  main fifteen-reader panel gives one estimate. Four stored summaries give
  the range on the ten readers available for every replay.}
  \label{fig:scaling-collapse}
\end{figure}

Fixed compression attenuates reader upgrades on both benchmarks. If
every reader gained the same amount, each upgrade would remain unchanged.
Figure~\ref{fig:scaling-collapse} places each reader's raw score against its
fixed-compression score. Instead, lower raw-scoring readers gain more, so the
reader-scaling curve flattens.

\paragraph{HotpotQA.}
Fixed RECOMP absorbs most of the visible reader upgrade. The upgrade between the
lowest- and highest-scoring raw readers falls from 31.8\pp{} under raw evidence to 7.8\pp{} under
compression, leaving 24.5\%. F1 gives the same retention estimate.
Table~\ref{tab:headline} reports the uncertainty intervals.

\paragraph{MuSiQue.}
MuSiQue supplies a second benchmark and compression family. On its main
fifteen-reader panel, gain falls as raw score rises under F1 ($r=-.951$) and EM
($r=-.827$) after correcting for shared row noise. Both intervals
remain below zero. The upgrade between its raw endpoints nearly vanishes, but its retention
intervals are wide (Table~\ref{tab:headline}).

To check whether attenuation persists across summary outputs, we replay four
artifacts on the same ten available readers. All attenuate their raw upgrade,
retaining 54--76\% across EM and F1. These ten readers exclude the
fifteen-reader endpoints, so the replay tests whether attenuation persists, not
its size.

\begin{table}[H]
\centering
\tablefont
\setlength{\tabcolsep}{4.2pt}
\begin{tabular}{@{}lllrrrr@{}}
\toprule
Benchmark & Panel & Metric & Readers & Raw & Comp. & Retained (\%) \\
\midrule
\benchswatch{benchhotpot}HotpotQA & main & EM & 20 & 31.8 & 7.8 & 24.5 $[-0.7,37.9]$ \\
\benchswatch{benchhotpot}HotpotQA & main & F1 & 20 & 34.7 & 8.5 & 24.5 $[5.0,34.4]$ \\
\benchswatch{benchmusique}MuSiQue & main & EM & 15 & 15.4 & 0.6 & 3.9 $[-35.0,85.7]$ \\
\benchswatch{benchmusique}MuSiQue & main & F1 & 15 & 18.7 & 0.1 & 0.8 $[-11.7,27.0]$ \\
\benchswatch{benchmusique}MuSiQue & four artifacts & EM & 10 & 13.8 & 7.4--8.6 & 53.6--62.3 \\
\benchswatch{benchmusique}MuSiQue & four artifacts & F1 & 10 & 15.6 & 8.4--11.9 & 54.0--76.4 \\
\bottomrule
\end{tabular}

\caption{\textbf{Both confirmatory benchmarks show reader-dependent
attenuation.} Raw and compressed upgrades are in pp and use the same pair
within each panel, chosen by raw score. Brackets are family-by-row intervals
for one stored artifact. MuSiQue ranges cover four correlated artifacts on the
same ten-reader panel.}
\label{tab:headline}
\end{table}

The effect extends beyond the endpoint pair. Across both benchmarks and
metrics, lower raw-scoring readers gain more after correcting for shared
sampling noise. The pattern remains after scaling each gain by the reader's
remaining accuracy headroom; every interval stays below zero. Compression
raises the panel means while flattening the reader curves.

\subsection{Attenuation varies across compressors}

Table~\ref{tab:policies} reports every panel with at least eight readers that
satisfies the shared-artifact rule. Rows are ordered by upgrade retention. Scores are native
EM for HotpotQA and MuSiQue, alias-max EM for NQ-Open and TriviaQA, and semantic
outcomes for LongMemEval. Its two rows share fifteen readers; every other row
uses its own verified panel.

\begin{table}[!htbp]
\centering
\tablefont
\setlength{\tabcolsep}{5pt}
\begin{tabular}{@{}llrrr@{}}
\toprule
Benchmark & Policy & Readers & Mean gain & \multicolumn{1}{c}{Retained (\%)} \\
\midrule
\benchswatch{benchhotpot}HotpotQA & EXIT & 20 & $+1.7$\pp{} & \retbar{benchhotpot}{83.6} \\
\benchswatch{benchmusique}MuSiQue & main summary replay & 10 & $+18.4$\pp{} & \retbar{benchmusique}{62.3} \\
\benchswatch{benchother}NQ-Open & EXIT & 8 & $-1.9$\pp{} & \retbar{benchother}{58.2} \\
\benchswatch{benchother}NQ-Open & RECOMP, raw fallback & 9 & $+1.4$\pp{} & \retbar{benchother}{47.3} \\
\benchswatch{benchother}LongMemEval & SIEVE & 15 & $+5.8$\pp{} & \retbar{benchother}{43.2} \\
\benchswatch{benchother}TriviaQA & shared summary & 13 & $-0.4$\pp{} & \retbar{benchother}{42.8} \\
\benchswatch{benchother}NQ-Open & Provence & 20 & $-2.3$\pp{} & \retbar{benchother}{40.3} \\
\benchswatch{benchhotpot}HotpotQA & RECOMP top-5 & 20 & $+7.4$\pp{} & \retbar{benchhotpot}{24.5} \\
\benchswatch{benchother}NQ-Open & shared summary & 20 & $-1.9$\pp{} & \retbar{benchother}{20.2} \\
\benchswatch{benchhotpot}HotpotQA & shared summary & 9 & $+13.6$\pp{} & \retbar{benchhotpot}{18.1} \\
\benchswatch{benchhotpot}HotpotQA & RECOMP abstractive & 11 & $-0.2$\pp{} & \retbar{benchhotpot}{13.8} \\
\benchswatch{benchother}NQ-Open & RECOMP top-5 & 20 & $-2.7$\pp{} & \retbar{benchother}{13.4} \\
\benchswatch{benchother}LongMemEval & shared summary & 15 & $+4.2$\pp{} & \retbar{benchother}{4.5} \\
\bottomrule
\end{tabular}
\caption{\textbf{Every listed point estimate attenuates the reader upgrade,
but by different amounts.} Rows are ordered by upgrade retention. MuSiQue uses
the main summary on the ten-reader replay panel. LongMemEval uses semantic
scores on the fifteen readers shared by its two panels.}
\label{tab:policies}
\end{table}

Across EM rows, retention is 13.4--83.6\%; LongMemEval's semantic score extends
the table to 4.5\%. F1 retention is 9.8--76.4\%
(Appendix~\ref{app:complementary}). This spread makes each deployed compressor's own
audit necessary.

\paragraph{Average gain and retention diverge.}
The shared HotpotQA summary raises the panel average by 13.6\pp{} but retains
18.1\% of the raw upgrade. By contrast, EXIT raises the average by only 1.7\pp{} but retains
83.6\%. An accuracy check therefore cannot show whether a compressor preserves
the reader comparison.

\FloatBarrier

\subsection{Controls bound alternative explanations}

Three controls test endpoint selection, unequal MuSiQue access, and whether any
evidence change produces the result.

First, selecting raw endpoints on one half of the questions and scoring them on
the other still shows attenuation. Second, giving every supplied MuSiQue
paragraph to three raw readers leaves 8--11\% EM retention and 11--15\% F1
retention across three summaries. Attenuation remains after equalizing candidate
access for those three readers.

Third, on a twelve-reader LongMemEval panel, dense retrieval changes 45.9\% of
the candidates. It still preserves 92.3\% of the raw upgrade and reverses none
of 43 separated pairs. Fixed SIEVE compression on the same panel retains
53.8\% and reverses two.
Compression therefore attenuates this matched panel more than a large retriever
change. The appendix reports the endpoint, retriever-swap, and
matched-input analyses separately
(Appendices~\ref{app:endpoint-sensitivity}, \ref{app:retriever-swap}, and
\ref{app:artifact-replay}).

\section{Compression Makes Reader Upgrades Harder to Resolve}
\label{sec:ranking}

\begin{figure}[H]
  \centering
  \includegraphics[width=\textwidth]{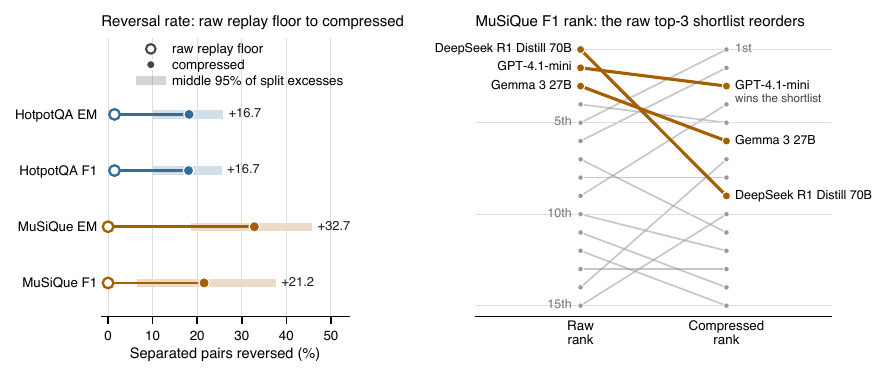}
  \caption{\textbf{Reader pairs change order more often under compressed
  evidence than in raw replays.} Left: open points show reversals after changing
  only the raw question half; filled points show reversals under compression.
  Labels give the median excess across 5,000 splits. Shaded bars cover the
  middle 95\% of split excesses. Right: the MuSiQue F1 shortlist winner
  changes under compressed evidence.}
  \label{fig:excess-flips}
\end{figure}

Smaller upgrades are harder to distinguish at the same question count.
They also cross zero more easily when the evaluated questions change. The raw
replay measures this background reversal rate. We then measure the additional
reversals under compressed evidence
(Figure~\ref{fig:excess-flips}).

Changing the raw question half reverses a median 1.5\% of eligible HotpotQA
pairs and none on MuSiQue. In contrast, compression adds 16.7\pp{} to the reversal rate on
HotpotQA under both metrics. The excess is 32.7\pp{} under MuSiQue EM and
21.2\pp{} under F1. None of the four shaded ranges includes zero. Other cutoffs
and tie rules leave every point estimate positive (Appendix~\ref{app:ranking-details}).

The practical loss is visible without extrapolation. On 250-question halves,
exact paired McNemar tests at $p<.05$ first identify pairs separated under raw
evidence. Only 41.7\% of those HotpotQA pairs remain significant in the same
direction under compression; MuSiQue retains 27.6\%. Most comparisons therefore
become unresolved at the same question count.

On MuSiQue, DeepSeek R1 Distill Llama 70B is the raw F1 winner, but GPT-4.1-mini
wins the compressed raw-top-three shortlist. EM, however, keeps its winner. The selected
reader therefore depends on the evidence policy in this example.

\paragraph{Smaller upgrades explain part of the excess.}
A smaller upgrade can flip by chance even when reader order stays the same. A
simulation that gives every reader one shared rescue rate and one shared damage
rate reproduces most of the HotpotQA
EM excess and about half of the MuSiQue EM excess. However, under EM, pairs reverse on
both question halves more often than in 99\% of HotpotQA simulations and in
every MuSiQue simulation. That rules out the shared-rate simulation. We added this check after the main
analysis, and MuSiQue F1 does not show the pattern
(Appendix~\ref{app:resolution-null}).

\FloatBarrier

\section{Rescue and Damage Coexist Inside the Average}
\label{sec:mechanism}

Average pipeline accuracy can rise while the visible reader upgrade shrinks.
Compression \emph{rescues} a raw-wrong answer when it becomes correct and
\emph{damages} a raw-correct answer when it becomes wrong. Different readers can
reach similar compressed averages through different paths. These transitions do
not identify why.

Both benchmarks show substantial movement in both directions. On HotpotQA,
19.1\% of all row-reader pairs are rescued. Of the pairs a reader answered
correctly under raw evidence, 35.7\% are damaged.

\begin{table}[H]
\centering
\tablefont
\setlength{\tabcolsep}{4pt}
\begin{tabular}{@{}lrrrr@{}}
\toprule
& \multicolumn{2}{c}{EM transitions} & \multicolumn{2}{c}{F1 score change} \\
\cmidrule(lr){2-3}\cmidrule(l){4-5}
Benchmark & Rescue & Damage & Mean increase & Mean decrease \\
\midrule
\benchswatch{benchhotpot}HotpotQA & 19.1 & 35.7 & 22.0 & 12.6 \\
\benchswatch{benchmusique}MuSiQue & 22.9 & 33.8 & 32.3 & 3.4 \\
\bottomrule
\end{tabular}

\caption{\textbf{Substantial rescue and damage coexist inside the same
average.} EM rescue is a percentage of all pairs and damage a percentage of
raw-correct pairs, so their difference is not the accuracy change. The F1 columns report average positive and negative score
changes in percentage points.}
\label{tab:transitions}
\end{table}

\section{Fixed-Artifact Reader Replay}
\label{sec:protocol}

For a declared reader change, run both readers on raw evidence and the same
compressed text. \emph{Fixed-Artifact Reader Replay} hashes the evidence and
reports scores, the surviving upgrade, changed answers, and uncertainty.
A pairwise choice requires those readers; a family-level claim requires a
family-spanning panel. For example, a three-reader shortcut fixed in advance, using the
lowest, middle, and highest raw scorers, caught only 1.9--8.5\% of the broader
panel's reversals
(Appendix~\ref{app:audit-validation}).

The \texttt{ragscale} workflow runs this audit from a configuration or from
existing paired scores. It compiles and hashes the compressed evidence once,
replays the declared readers under raw and compressed evidence, and reports
retention, reversals, rescue, damage, uncertainty, and resource use. As a
software check, it reproduces every HotpotQA measure and rejects changed
hashes. This check adds no benchmark evidence
(Appendix~\ref{app:ragscale-validation}).

\section{Limitations}
\label{sec:limitations}

Our claims cover the observed readers, prompts, evidence policies, and stored
artifacts. The panels are nonrandom and exclude readers below 7B. We lack
production A/B-test logs, cross-prompt analysis, and causal mechanism tests. The
MuSiQue equal-access check covers only three readers. Endpoint retention is
panel-specific and sometimes imprecise. Resolution and
cost cover HotpotQA and MuSiQue; cost estimates project beyond 500
questions and exclude compressor calls.
A predictor fixed in advance from raw reader spacing and compressor class
performed worse than raw spacing alone. Direct panel audits remain necessary
(Appendix~\ref{app:predictive-audit}).

\section{Conclusion}
\label{sec:conclusion}

A useful system component need not be a neutral measurement layer. One
in-domain RECOMP artifact improved several HotpotQA pipelines while shrinking
the upgrade between the lowest- and highest-scoring raw readers from 31.8\pp{}
to 7.8\pp{}. Compression rescued and damaged answers at different rates for
different readers, so average accuracy rose while the reader upgrade shrank.

Across EM panels, retention ranged from 13.4\% to 83.6\%. Evaluations should report raw
results and the surviving upgrade under the same compressed text; the released
dataset and \texttt{ragscale} workflow reproduce this audit.

\section*{Ethics Statement}

Reader rankings can influence procurement and deployment. Reporting raw and
compressed results makes their dependence on the evidence policy visible. The
released artifacts use separately cited public benchmarks and contain no newly
collected personal data.

\section*{AI Use Statement}
Generative AI tools assisted with code, figures, and language. The authors
chose the framing, verified every statistic against frozen artifacts, and take responsibility for the paper and the release.

\section*{Reproducibility Statement}
The appendix gives provenance, reader panels, scoring, estimators, resampling,
exclusions, and reproduction commands.
\ifpreprint
The \href{\codeurl}{repository} provides \texttt{ragscale}, its tests, worked
examples, and the interaction matrix of stored binary outcomes.
\fi

\bibliography{references}
\bibliographystyle{iclr2027_conference}

\appendix
\section{Replication Material}
\label{app:replication}

\subsection{Artifact provenance}

The public benchmark questions and references come from HotpotQA, MuSiQue,
LongMemEval, NQ-Open, and TriviaQA
\citep{yang2018hotpotqa,trivedi2022musique,wu2025longmemeval,
kwiatkowski2019natural,joshi-etal-2017-triviaqa}. We construct the experimental layer: fixed slices, candidate pools,
compressed evidence, reader generations, judged outputs, scores,
provenance records, and the 176,864-row interaction matrix.

For benchmark $d$, method $c$, and row $i$, the provenance key contains the
source row identifier, candidate-pool identifier, compression method and
configuration, compressed artifact, and artifact hash. A reader joins a panel
only if every row key matches the panel footprint. We intersect raw and
compressed row identifiers before scoring. No missing score is imputed.

The confirmatory HotpotQA artifact is one trained extractive RECOMP output per
row \citep{xu2024recomp}. The MuSiQue artifact is the largest exact shared-summary footprint. Its
fifteen readers receive byte-identical summaries. LongMemEval SIEVE is compiled
once per row for the twenty-reader panel. The LongMemEval summary replication
uses one shared GPT-4.1-mini summary cache across seventeen readers.

The later TriviaQA robustness study replays one 500-row Qwen3 8B summary cache
across thirteen readers. Its post-freeze cache and generations are not rows in
the 176,864-row primary matrix \citep{joshi-etal-2017-triviaqa}.

The released interaction matrix contains 176,864 rows and excludes a larger
sibling workspace. Release metadata records panel membership, benchmark row
counts, and per-file hashes. It also distinguishes the upstream benchmarks
from our generated interaction data.

\subsection{Reader inventory}
\label{app:readers}

The twenty-reader HotpotQA panel contains OLMo 3.1 32B Instruct, Claude 3.5 
Haiku, Seed 2.0 Mini, Command R, DeepSeek R1 Distill Llama 70B, Gemma 3 12B and
27B, Llama 3.1 8B and 70B, Llama 3.3 70B, Llama 4 Scout, Phi 4, GPT-4.1-mini,
Qwen 2.5 7B and 72B Instruct, Qwen3 8B, 14B, and 32B, Grok 4.1 Fast, and GLM-4
32B. These readers form eleven family clusters.

MuSiQue uses the same catalog except Gemma 3 12B, both Llama 3.1 checkpoints,
and both Qwen 2.5 checkpoints. Its fifteen readers form eleven family clusters.
The LongMemEval SIEVE panel contains the same twenty-reader catalog. The shared
summary footprint contains seventeen readers and adds Qwen3.6 27B and MiMo
v2.5 while omitting OLMo 3.1 32B, Llama 3.1 8B, Llama 4 Scout, Grok 4.1 Fast,
and GLM-4 32B. Hosted model names identify the versions represented in the
cached artifact. Current provider endpoints may differ.

TriviaQA adds Qwen3.5 9B, 27B, and 397B-A17B to Command R, Gemma 3 12B and
27B, Llama 3.1 8B and 70B, Llama 4 Scout, Phi-4, GPT-4.1-mini, Qwen 2.5 7B,
and Qwen3 8B. These readers form six family clusters.

\subsection{Answer scoring}
\label{app:scoring}

The scorer lowercases text, removes English articles, removes punctuation, and
collapses whitespace according to each benchmark's native script. HotpotQA
special-cases yes, no, and no-answer references \citep{yang2018hotpotqa}.
MuSiQue maximizes EM and token F1 over supplied aliases
\citep{trivedi2022musique}. Token F1 is the multiset overlap between normalized
prediction and reference tokens. Empty normalized strings score one only when
both prediction and reference are empty.

TriviaQA uses the same normalized alias-max EM and token-F1 definitions over
its supplied answer aliases \citep{joshi-etal-2017-triviaqa}. No LLM judge or
answer extraction is applied.

LongMemEval uses the pinned DeepSeek V3 endpoint
\nolinkurl{deepseek/deepseek-chat-v3-0324} at temperature zero with a 10-token
output limit. We follow the question-type-specific binary prompts released by
\citet{wu2025longmemeval}. The general rubric supplies the question,
reference, and model response. It asks for ``yes'' when the response contains
the correct answer, is equivalent to it, or contains all required intermediate
steps. A response containing only a required subset receives ``no.'' Temporal
questions tolerate off-by-one duration counts. Knowledge-update questions
accept earlier information when the required updated answer is also present.
Preference questions require correct recall and use of the user's information.
Abstention questions require the response to identify the question as
unanswerable. A response containing ``yes'' maps to score 2 and correct;
otherwise it maps to score 0 and incorrect. The release contains all five
templates verbatim.

NQ-Open uses the same alias-max lexical interface but is not scored with
Google's document-level evaluator \citep{kwiatkowski2019natural}. The source reference for \texttt{test2104}
is ``)'', which normalizes to empty. We exclude the row before analysis and
retain the malformed source value for provenance. The resulting lexical
panels contain 323 rows.

\subsection{Paired errors-in-variables correction}

Compression gain already contains the raw score, so their observed covariance
also contains shared row-sampling noise. The paired errors-in-variables (EIV)
correction removes that sampling contribution before estimating their
cross-reader correlation.

Let $R$ and $D$ be reader-by-row matrices for raw scores and paired gains. Let
$M$ be the observed cross-reader covariance of their row means. Because all
readers answer the same rows, sampling errors are correlated across readers.
For centered row matrices $R_c$ and $D_c$, we estimate
\begin{align}
  \Sigma_R &= \frac{R_cR_c^\top}{n(n-1)}, &
  \Sigma_D &= \frac{D_cD_c^\top}{n(n-1)}, \\
  \Sigma_{RD} &= \frac{R_cD_c^\top}{n(n-1)}.
\end{align}
With reader-centering matrix $H=I-\mathbf{1}\mathbf{1}^\top/m$, each sampling
moment is $\operatorname{tr}(H\Sigma)/(m-1)$. Subtracting the three projected
moments from $M$ gives $M^*$. The reported correlation is
\begin{equation}
 r_{\mathrm{EIV}} = \frac{M^*_{12}}
 {\sqrt{M^*_{11}M^*_{22}}}.
\end{equation}
We recompute the full nonlinear estimate in every bootstrap draw. The disjoint
row diagnostic estimates raw score on one row half and gain on the other, then
repeats the split 5,000 times.

Moment subtraction can produce a non-PSD matrix. We therefore
store $M^*_{11}$, $M^*_{22}$, $M^*_{12}$, its minimum eigenvalue, the unbounded
correlation, and the corrected slope. For continuity with the prespecified analysis,
the displayed bounded correlation clips finite values to $[-1,1]$. We also
report the correlation after truncating negative eigenvalues of $M^*$ to zero.
Every bootstrap records nonpositive corrected variances, non-PSD matrices,
nonfinite correlations, and finite values outside $[-1,1]$. The generated
report exposes every count; no boundary hit is silently replaced.

All four confirmatory point estimates are admissible. Across 5,000
family-by-row draws, the unbounded correlation falls outside $[-1,1]$ in
9 HotpotQA EM, 8 HotpotQA F1, 24 MuSiQue EM, and 129 MuSiQue F1 draws. Only one
HotpotQA EM draw has a nonpositive corrected variance. The PSD projection is
invoked in 10, 8, 24, and 129 draws, respectively. The highest invocation rate
is therefore 2.58\%; the confirmatory gates do not depend on a clipped
correlation.

\subsection{Ranking replay details}
\label{app:ranking-details}

This replay asks whether compression reverses reader pairs more often than a
repeated raw-evidence evaluation. Excess flips are the difference between
those two reversal rates.

Each of 5,000 iterations divides rows into selection and scoring halves. A pair
is eligible when its raw upgrade magnitude is at least 5\pp{} on the
selection half. We orient the pair by that sign, then record whether the sign
reverses on the scoring half under raw and compressed evidence. The difference
between compressed and raw reversal rates is computed within the iteration.
The reported point and interval are its median and 2.5th and 97.5th
percentiles. A scoring-half tie is not a reversal. Empty eligible-pair draws
are omitted. Each retained split contributes one eligible-pair proportion
before aggregation.

The EM sensitivity also requires a significant paired McNemar test on
the selection half. We use the exact two-sided binomial test at $p < .05$, with
no continuity or mid-$p$ adjustment. This filter leaves median eligible-pair counts of 123 on
HotpotQA and 56 on MuSiQue. It yields 15.2\pp{} excess flips
$[8.0,22.8]$ and 32.7\pp{} $[18.9,46.4]$, respectively.

\paragraph{Cutoff and tie sensitivity.} Changing the prespecified 5\pp{}
raw-upgrade cutoff to 2\pp{} or 10\pp{} leaves every point estimate positive.
At 10\pp{}, excess flips are 11.5\pp{} $[4.7,19.6]$ and 10.0\pp{}
$[3.7,17.4]$ on HotpotQA for EM and F1. On MuSiQue they are 37.9\pp{}
$[16.7,60.9]$ and 15.2\pp{} $[0.0,39.3]$. Only the MuSiQue F1 interval reaches
zero, with a median of 31 eligible pairs. At
2\pp{} the estimates are 18.0, 17.2, 33.3, and 26.5\pp{}, all with
intervals above zero. Exact scoring-half ties occur in at most 4.6\% of
eligible pairs. Excluding tied pairs, or counting every raw tie as an adverse
raw reversal, moves each estimate by less than 2\pp{}.

\subsection{Order-preserving null and resolution cost}
\label{app:resolution-null}

This check replays the stored scores to ask whether the excess flips in
Section~\ref{sec:ranking} exceed what uniform upgrade attenuation alone
produces, and what lost resolution costs. It uses the same rows, readers, and split
procedure as the replay and does not change its estimates. We use a seed
separate from the primary replay. We added the diagnostic after identifying the
uniform-contraction explanation; the primary attenuation analysis was already
fixed.

\paragraph{Nulls.} The \emph{independent-uniform} null keeps every raw
outcome, rescues each raw-wrong cell with the pooled rescue rate, and damages
each raw-correct cell with the pooled damage rate, independently across cells.
Expected compressed accuracy is then an increasing affine function of raw
accuracy, so every expected ordering is preserved and every reader upgrade is
multiplied by the same positive factor. It is defined for EM only. The \emph{row-exchangeable} null
permutes compressed outcomes within each row among readers that share the same
raw outcome. This preserves each row's rescue and damage counts but removes
any reader-specific response. It collapses compressed upgrades toward zero and
provides a second reader-blind reference. Each null is drawn 200 times
and replayed over 250 row splits; the observed statistic uses 5,000 splits.

\paragraph{Both-half reversals.} A raw-separated pair counts when its
compressed ordering opposes the raw ordering on both halves. This repeated
sign is stricter than a reversal on the scoring half alone, but it does not
require the compressed upgrades to be significantly different from zero. 

\begin{table}[H]
\centering
\tablefont
\setlength{\tabcolsep}{4pt}
\setlength{\tabcolsep}{3pt}
\begin{tabular}{@{}llrrrrr@{}}
\toprule
& & \multicolumn{3}{c}{Excess flips} & \multicolumn{2}{c}{Both-halves reversal} \\
\cmidrule(lr){3-5}\cmidrule(l){6-7}
Cell & Metric & Observed & Ind.-unif.\ null & Row-exch.\ null & Observed & Ind.-unif.\ null \\
\midrule
HotpotQA & EM & 16.7 [8.7, 24.4] & 12.5 [6.8, 20.5] & 23.5 [14.3, 33.8] & 11.9 [6.7, 17.2] & 2.4 [0.0, 7.2] \\
HotpotQA & F1 & 16.7 [8.5, 24.1] & -- & 25.8 [16.8, 37.4] & 11.4 [7.0, 15.8] & -- \\
MuSiQue & EM & 32.7 [18.6, 45.8] & 16.7 [7.8, 30.5] & 38.6 [23.2, 56.6] & 21.3 [12.5, 29.5] & 1.9 [0.0, 10.6] \\
MuSiQue & F1 & 21.2 [6.5, 37.7] & -- & 46.3 [33.3, 62.2] & 6.2 [1.5, 12.5] & -- \\
\bottomrule
\end{tabular}
\caption{\textbf{Uniform contraction explains much of the excess-flip rate.}
Values are percentages of raw-separated pairs.
Observed excess flips are the \ref{sec:ranking} estimates.
Both-halves rates and null columns come from the separately seeded
calibration run, whose observed excess flips agree within 0.2\pp{}. Observed
brackets are 2.5--97.5 percentiles over 5,000 splits; null brackets are the
same percentiles over 200 null draws. The observed both-halves rate
exceeds the independent-uniform null in 0.99 (HotpotQA) and 1.00 (MuSiQue) of
draws under EM. This comparison rejects only the simple uniform reference;
other order-preserving processes remain possible.}
\label{tab:resolution-null}
\end{table}

Per-reader transition rates are close to uniform. Rescue rates have a
cross-reader standard deviation of 0.027 on HotpotQA and 0.029 on MuSiQue;
damage rates 0.064 and 0.083. A chi-square test rejects homogeneity for
HotpotQA damage ($p=8\times10^{-4}$) and MuSiQue rescue ($p=.009$) only.
Real transitions are also correlated across readers within a row, which is why
the row-exchangeable null, with random assignment within each raw-outcome
group, produces more flips than the data.

\paragraph{Resolution cost.} Section~\ref{sec:ranking} showed that
compression attenuates reader upgrades. A smaller upgrade needs more questions to
resolve. Compressed prompts are also shorter, so each question costs less.
This paragraph asks which effect wins when compressed evidence stands in for
the raw comparison. It applies only to that proxy use. A team that deploys the
compressor cares about the smaller compressed upgrade itself. The projection
addresses measurement of the raw comparison.

We measure the trade-off two ways. The first is measured directly. On
250-question halves with exact McNemar tests, 41.7\% [30.8, 52.1] of HotpotQA
pairs that are significantly separated on raw evidence remain significantly 
separated in the same direction under compression. Another 2.5\% [0.0, 7.1]
become significant in the opposite direction. On MuSiQue the values are 
27.6\% [18.2, 38.2] and 3.5\% [0.0, 11.9]. At equal question counts,
compression therefore leaves most raw-resolved comparisons unresolved.

The second projects how many questions would restore the lost resolution. For
readers $a$ and $b$, let $D$ be each question's contribution to their reader
upgrade. The number of questions needed to recover the raw signal-to-noise
ratio scales as
$\operatorname{Var}(D^{\comp})/\operatorname{Var}(D^{\raw})/\rho^2$, where
$\rho$ is the pair's retention. This projection assumes independent questions
and a normal approximation. It excludes compiler cost, output tokens, and
pricing.
Table~\ref{tab:resolution-cost} reports the median over all pairs with a raw
upgrade of at least 5\pp{}. It also reports a cross-fit version that selects
pairs on one half and estimates on the other, and the product with the pair's
mean reader prompt-token ratio. Pairs with non-positive observed retention have no finite
ratio under this projection and are excluded from the medians. The endpoint
MuSiQue pairs have retention near zero and give ratios in the hundreds or
thousands; we do not report them as estimates.

The cross-fit projection needs a median of 5.9 times as many HotpotQA questions
under EM. Separately, the full-sample pair estimates combine each pair's
sample-size ratio with its prompt-token ratio. Compressed prompts use 28\% of
the raw reader tokens, and the resulting token-cost ratio has a median of 1.85
with an interquartile range of 0.97 to 5.18. Compression makes each question
cheaper, while the median estimated cost of resolving the raw comparison is
higher. The two summaries use different estimators and should not be
multiplied together.

\begin{table}[H]
\centering
\tablefont
\setlength{\tabcolsep}{4pt}
\begin{tabular}{@{}llrrrrr@{}}
\toprule
Cell & Metric & Pairs & Unresolv. & Cross-fit $n$ ratio & Mean token ratio & Full-sample token-cost ratio [Q1, Q3] \\
\midrule
HotpotQA & EM & 131 & 21 & 5.93 [4.06, 8.95] & 0.284 & 1.85 [0.97, 5.18] \\
HotpotQA & F1 & 127 & 21 & 7.22 [4.99, 10.79] & 0.284 & 2.33 [1.05, 6.01] \\
MuSiQue & EM & 54 & 18 & 3.36 [1.72, 8.76] & 0.314 & 1.24 [0.50, 5.16] \\
MuSiQue & F1 & 67 & 12 & 10.10 [5.09, 23.46] & 0.314 & 5.17 [1.06, 22.56] \\
\bottomrule
\end{tabular}
\caption{\textbf{Shorter prompts do not guarantee a cheaper reader
comparison.} Cross-fit brackets are
2.5--97.5 percentiles of the per-split median over 5,000 splits. Token ratio is
the mean compressed-to-raw reader prompt-token ratio across the panel, measured
from the stored runs. Token-cost ratio is the full-sample median of the
per-pair $n$ ratio times the pair's token ratio.}
\label{tab:resolution-cost}
\end{table}

\subsection{Minimum-audit validation}
\label{app:audit-validation}

Before inspecting validation outputs, we froze a catalog-informed three-reader
screen. On each selection half, it chooses the lowest and highest raw-scoring
readers. The middle reader is closest to their score midpoint, with preference
for a different family. Compressed scores never affect reader selection. The
screen expands when endpoint or admissible adjacent-pair retention is below
.75, median excess flips exceed 5\pp{}, a pair reverses, or a required estimate
is missing. A screen-negative result is not labeled safe.

We compare this screen with the broader panel on the same 5,000 held-out row
splits. The cached validation covers 24 metric cells from thirteen
provenance-valid panels: the native-score confirmatory cells, HotpotQA and
NQ-Open sensitivities, two LongMemEval replications, and the later TriviaQA
panel. Metrics and methods within one benchmark remain correlated views.

\begin{table}[h]
\centering
\tablefont
\setlength{\tabcolsep}{3.5pt}
\begin{adjustbox}{max width=\columnwidth}
\begin{tabular}{@{}llrrr@{}}
\toprule
Benchmark & Metric & False triplets & Triplet identities & Reversals visible \\
\midrule
\benchswatch{benchhotpot}HotpotQA & EM & 1 & 1,278 & 3.1\% \\
\benchswatch{benchhotpot}HotpotQA & F1 & 2 & 1,025 & 1.9\% \\
\benchswatch{benchmusique}MuSiQue & EM & 34 & 451 & 7.0\% \\
\benchswatch{benchmusique}MuSiQue & F1 & 9 & 376 & 8.5\% \\
\bottomrule
\end{tabular}
\end{adjustbox}
\caption{\textbf{The prespecified three-reader validation fails its family-choice
gate.}
False triplets are ordered weak--middle--strong identities whose aggregate
screen is negative while the broader panel requires expansion. Reversals
visible is the share of broader-panel compressed reversals whose two readers
both occur in the primary triplet.}
\label{tab:audit-validation}
\end{table}

The catalog-informed primary triplet calls for expansion in all four
confirmatory cells, so that particular triplet never gives false reassurance.
The broader validation still fails the prespecified family-choice gate. Alternative
family-distinct triplets can miss the need to expand, and the primary triplet
observes few broader-panel reversals. Endpoint retention is also reproduced
partly by construction.
Sensitivity checks at retention .50 and 1.00 and excess flips 0 and 10\pp{} do
not change this conclusion. They do not replace the prespecified .75 and 5\pp{} rule.

Without an existing raw catalog, a three-reader illustration requires three
raw and three compressed reader runs. With a raw catalog, it adds three
compressed runs. Because the screen fails, these costs do not replace a
broader family-spanning replay. Compilation, hashing, retries, provider
variation, judge replay, and new artifact draws remain separate costs.

\section{Complementary Results}
\label{app:complementary}

\subsection{Endpoint selection and panel floor}
\label{app:endpoint-sensitivity}

The point estimator selects the raw weakest and strongest readers on all rows.
To expose extremum-selection sensitivity, we select them on one row half and
score the same pair on the other. Both orientations are repeated across 5,000
permutations. Table~\ref{tab:endpoint-sensitivity} reports the median and
2.5th--97.5th percentile range. These split-sensitivity ranges do not estimate
population uncertainty or correct a population extremum.
All 10,000 held-out orientations satisfy the 5\pp{} rule. Endpoint identities
do turn over. HotpotQA selects 14 EM pairs and 20 F1 pairs; their modal shares
are 39.0\% and 33.7\%. MuSiQue selects 16 and 11 pairs, with modal shares of
73.6\% and 54.6\%. LongMemEval selects 9 SIEVE pairs and 7 summary pairs, with
modal shares of 57.7\% and 66.6\%.

\begin{table}[h]
\centering
\tablefont
\setlength{\tabcolsep}{3.5pt}
\begin{adjustbox}{max width=\columnwidth}
\begin{tabular}{@{}llrrr@{}}
\toprule
Panel & Metric & Median $\rho$ & Split range & Valid \\
\midrule
\benchswatch{benchhotpot}HotpotQA & EM & 0.241 & [0.101, 0.372] & 10000/10000 \\
\benchswatch{benchhotpot}HotpotQA & F1 & 0.215 & [0.080, 0.344] & 10000/10000 \\
\benchswatch{benchmusique}MuSiQue & EM & 0.000 & [-0.333, 0.233] & 10000/10000 \\
\benchswatch{benchmusique}MuSiQue & F1 & 0.025 & [-0.131, 0.193] & 10000/10000 \\
\benchswatch{benchother}LongMemEval SIEVE & Semantic & 0.576 & [0.437, 0.750] & 10000/10000 \\
\benchswatch{benchother}LongMemEval Summary & Semantic & 0.055 & [-0.119, 0.255] & 10000/10000 \\
\bottomrule
\end{tabular}

\end{adjustbox}
\caption{Held-out endpoint retention is stable on HotpotQA and LongMemEval.
MuSiQue remains strongly attenuated but its endpoint estimate is less precise.
Valid gives ratios meeting the 5\pp{} held-out raw-upgrade rule.}
\label{tab:endpoint-sensitivity}
\end{table}

Table~\ref{tab:floor-diagnostic} compares the upgrade between the lowest- and
highest-scoring raw readers with the full reader range. MuSiQue compression raises the panel while leaving a broad
range of reader scores.

\begin{table}[h]
\centering
\tablefont
\setlength{\tabcolsep}{3pt}
\begin{adjustbox}{max width=\columnwidth}
\begin{tabular}{@{}llrrrr@{}}
\toprule
Panel & Metric & Raw mean & Comp. mean & Raw range & Comp. range \\
\midrule
\benchswatch{benchhotpot}HotpotQA & EM & 0.330 & 0.404 & 31.8 & 11.8 \\
\benchswatch{benchhotpot}HotpotQA & F1 & 0.461 & 0.555 & 34.7 & 10.6 \\
\benchswatch{benchmusique}MuSiQue & EM & 0.091 & 0.289 & 15.4 & 11.8 \\
\benchswatch{benchmusique}MuSiQue & F1 & 0.136 & 0.425 & 18.7 & 9.0 \\
\benchswatch{benchother}LongMemEval SIEVE & Semantic & 0.451 & 0.519 & 28.8 & 17.4 \\
\benchswatch{benchother}LongMemEval Summary & Semantic & 0.481 & 0.516 & 22.2 & 12.4 \\
\bottomrule
\end{tabular}

\end{adjustbox}
\caption{\textbf{Compression raises the MuSiQue panel and leaves a broad reader
range.} Values are unweighted panel means and full reader ranges.
Range columns are percentage points; the compressed range is not endpoint
retention.}
\label{tab:floor-diagnostic}
\end{table}

The family-by-row endpoint bootstrap discards one of 5,000 draws for HotpotQA
EM, none for HotpotQA F1 or MuSiQue EM, and one for MuSiQue F1. Including every
nonzero denominator changes no reported endpoint at three decimals. The SIEVE
replication discards none; the clean LongMemEval summary discards one.

\subsection{Correlated method and boundary cells}

Tables~\ref{tab:hotpot-sensitivities} and~\ref{tab:nq-sensitivities} report the
remaining provenance-valid answer
metric panels. Rows within a benchmark are correlated sensitivity analyses.
They broaden sensitivity coverage while leaving the number of independent
replications unchanged. We report cell-wise resampling intervals and count each
benchmark once. We apply no multiple-comparisons correction. The
NQ-Open rows use lexical scores and remain boundary evidence even when their
intervals exclude the raw replay floor. These settings use RECOMP, EXIT, and
Provence \citep{xu2024recomp,exit2025,provence2025}.

\begin{table}[t]
\centering
\tablefont
\setlength{\tabcolsep}{3.5pt}
\begin{adjustbox}{max width=\columnwidth}
\begin{tabular}{@{}lrrr@{}}
\toprule
Setting & $m$ & Retention EM / F1 & Excess EM / F1 \\
\midrule
EXIT extractive & 20 & $.836 / .653$ & $4.4 / 4.6$ \\
RECOMP abstractive & 11 & $.138 / .172$ & $31.0 / 32.6$ \\
Shared summary & 9 & $.181 / .098$ & $12.5 / 16.7$ \\
\bottomrule
\end{tabular}%
\end{adjustbox}
\caption{HotpotQA sensitivity settings vary in magnitude. Excess values are
percentage points. These correlated cells do not count as new benchmarks.}
\label{tab:hotpot-sensitivities}
\end{table}

\begin{table}[t]
\centering
\tablefont
\setlength{\tabcolsep}{3.5pt}
\begin{adjustbox}{max width=\columnwidth}
\begin{tabular}{@{}lrrr@{}}
\toprule
Setting & $m$ & Retention EM / F1 & Excess EM / F1 \\
\midrule
RECOMP top-5 & 20 & $.134 / .312$ & $13.9 / 16.8$ \\
Shared summary & 20 & $.202 / .308$ & $10.4 / 13.8$ \\
EXIT extractive & 8 & $.582 / .527$ & $0.0 / 0.0$ \\
Provence \citep{provence2025} & 20 & $.403 / .479$ & $5.6 / 4.3$ \\
RECOMP raw fallback & 9 & $.473 / .419$ & $10.7 / 4.3$ \\
\bottomrule
\end{tabular}%
\end{adjustbox}
\caption{NQ-Open lexical boundary settings. Retention is a ratio; excess
values are percentage points. These are not native NQ scores.}
\label{tab:nq-sensitivities}
\end{table}

\subsection{TriviaQA third-benchmark robustness}
\label{app:triviaqa}

We first froze a three-reader TriviaQA test: 500 \texttt{rc.wikipedia}
validation questions, BM25 top-20 evidence, a generic Qwen3 8B compressor,
provider routes, and a 5\pp{} gate \citep{joshi-etal-2017-triviaqa}. F1 retained
0.376 of a 7.0\pp{} raw upgrade $[0.182,0.741]$. EM retained 1.031 of a
6.4\pp{} upgrade and did not attenuate. We preserve this metric disagreement.

The F1 result met the predeclared expansion trigger. Score-blind execution
checks then admitted ten of sixteen candidates without replacement. Adding the
three completed Qwen3.5 readers produced the panel below. Because the result
triggered its own expansion, the final panel has robustness status. It does not
add a confirmatory benchmark.

\begin{table}[h]
\centering
\tablefont
\setlength{\tabcolsep}{3pt}
\begin{adjustbox}{max width=\columnwidth}
\begin{tabular}{@{}lrrrrr@{}}
\toprule
Metric & Raw upgrade & Comp. upgrade & Retention & Family-by-row interval & Excess flips \\
\midrule
EM & 53.8 & 23.0 & .428 & $[.000,.587]$ & 16.3 $[4.3,25.0]$ \\
F1 & 39.7 & 15.5 & .391 & $[-.001,.520]$ & 17.6 $[6.8,28.0]$ \\
\bottomrule
\end{tabular}%
\end{adjustbox}
\caption{\textbf{The expanded TriviaQA panel attenuates reader upgrades under
both native metrics.} Reader upgrades and excess flips are percentage points.
Retention intervals resample reader families and rows; excess-flip intervals
are panel-conditional split-stability ranges.}
\label{tab:triviaqa-robustness}
\end{table}

Errors-in-variables (EIV) correlations are $-0.824$ $[-0.987,-0.116]$ for EM and $-0.871$
$[-0.989,-0.652]$ for F1. Every split selects Phi-4 and Llama 3.1 70B as the
endpoints. All twenty-six run artifacts pass the 500-row protocol audit. The
study covers one summary draw on a sampled validation panel; the answer-hidden
test split remains outside its scope.

\subsection{Binary semantic-outcome replication}

The LongMemEval semantic-judge analysis predates the lexical scoring
pass but uses the same paired errors-in-variables (EIV) estimator.
Table~\ref{tab:semantic-replication} collects the three diagnostics. Both
compressors show a negative raw-score--gain correlation and positive excess
flips. On the fifteen-reader intersection, the raw judged outcomes match in
all 7,500 shared reader-row cells, and both policies use the same Qwen 2.5
7B--Seed 2.0 Mini endpoints.

\begin{table}[!t]
\centering
\tablefont
\setlength{\tabcolsep}{3.5pt}
\begin{adjustbox}{max width=\columnwidth}
\begin{tabular}{@{}lrrr@{}}
\toprule
Compressor & EIV $r$ & Retention & Excess flips (pp) \\
\midrule
SIEVE & $-.916\;[-.996,-.743]$ & .432 & $6.7\;[1.5,13.0]$ \\
Shared summary & $-.801\;[-.958,-.371]$ & .045 & $35.3\;[18.7,48.1]$ \\
\bottomrule
\end{tabular}
\end{adjustbox}
\caption{\textbf{Both LongMemEval compressors attenuate the reader upgrade.}
Retention divides the visible compressed upgrade by the raw upgrade on the
fifteen-reader intersection. EIV intervals resample families and rows. Excess
flips are compressed minus raw reversal rates; their intervals measure
panel-conditional split stability.}
\label{tab:semantic-replication}
\end{table}

\begin{table}[!t]
\centering
\tablefont
\setlength{\tabcolsep}{4pt}
\begin{adjustbox}{max width=\columnwidth}
\begin{tabular}{@{}lrr@{}}
\toprule
Measure & BM25 top-20 & Dense top-20 \\
\midrule
Candidate turnover (\%) & 0.0 & 45.9 \\ 
Paired EIV $r$ & $-0.913$ $[-1.000,-0.721]$ & $-0.977$ $[-1.000,-0.759]$ \\ 
Raw reader upgrade (pp) & 27.8 & 25.6 \\
Visible compressed upgrade (pp) & 15.0 & 13.5 \\
Upgrade retention & $0.538$ $[0.203,0.731]$ & $0.525$ $[0.232,0.762]$ \\ 
Excess flips (pp) & $8.7$ $[0.0,18.2]$ & $4.5$ $[-4.3,15.0]$ \\ 
\bottomrule
\end{tabular}

\end{adjustbox}
\caption{\textbf{SIEVE attenuation changes little after the retriever changes.}
This matched LongMemEval control compares BM25 and dense retrieval. Brackets
are family-by-row intervals except for excess flips, whose brackets are
panel-conditional split-stability intervals.}
\label{tab:retrieval-control}
\end{table}

Table~\ref{tab:retrieval-control} restricts BM25 and dense retrieval to the
same twelve readers and 468 judged rows. Dense retrieval replaces 45.9\% of
the BM25 candidates, yet SIEVE retention changes little: 0.538 under BM25 and
0.525 under dense retrieval. Both paired EIV intervals exclude zero. The dense
excess-flip interval includes zero, leaving the ranking-instability result
inconclusive in this control.

\subsection{Is compression special, or is any evidence change enough?}
\label{app:retriever-swap}

The control above varies the retriever and then compresses. A reviewer may
reasonably ask the prior question: does a non-compressing change to the shared
evidence attenuate reader upgrades on its own? The same cached panel answers it.
Swapping BM25 for dense retrieval replaces 45.9\% of the candidate pool while
leaving every retained passage unedited. This gives a large, non-compressing
input change.

\begin{table}[!t]
\centering
\tablefont
\setlength{\tabcolsep}{4pt}
\begin{adjustbox}{max width=\columnwidth}
\begin{tabular}{@{}lrrr@{}}
\toprule
Intervention & Retention & Panel mean (pp) & Excess flips (pp) \\
\midrule
Retriever swap (BM25 $\to$ dense) & $0.923$ $[0.543,1.129]$ & $-3.6$ & $2.1$ $[-6.2,8.2]$ \\ 
Compression (BM25 raw $\to$ SIEVE) & $0.538$ $[0.203,0.731]$ & $+6.3$ & $8.7$ $[0.0,18.2]$ \\ 
Compression (dense raw $\to$ SIEVE) & $0.525$ $[0.232,0.762]$ & $+5.1$ & $4.5$ $[-4.3,15.0]$ \\ 
\bottomrule
\end{tabular}

\end{adjustbox}
\caption{\textbf{The retriever swap preserves more of the reader upgrade than
fixed compression.} Both interventions use the same twelve readers and 468
rows. Endpoints are selected under each intervention's reference condition.
Retention brackets are family-by-row intervals; excess-flip brackets are
panel-conditional split-stability intervals. Panel mean is the mean change in
reader accuracy.}
\label{tab:retriever-swap}
\end{table}

Table~\ref{tab:retriever-swap} separates the two interventions. The swap
preserves the reader upgrade, retaining $0.923$ $[0.543, 1.129]$ of it. It also
preserves the ordering: no eligible pair reverses, and the excess-flip rate of
$2.1$\pp{} is positive in 0.52 of split draws, as expected under the null.
Compression halves the same upgrade, to $0.538$ under BM25 and $0.525$ under
dense retrieval. Its excess flips are positive in 0.94 of draws. A paired
bootstrap that selects endpoints once per draw puts the retention difference
at $0.372$ $[0.119, 0.623]$.

The swap has the larger accuracy cost: it lowers panel accuracy by $3.6$pp,
while compression adds $6.3$pp. Compression nevertheless causes the larger
loss of reader-upgrade resolution. This contrast separates compression from a
generic evidence-change explanation on this panel.

Two cautions limit this control. First, the swap contrast has a wide EIV
interval that crosses zero. Preservation by retriever swaps therefore remains
unestablished. Second, only 43
reader pairs are eligible, so the excess-flip percentiles move in steps of
$1/43$. The BM25 compression lower bound sits at zero and reaches
$-2.0$\pp{} under one of five checked replay seeds. Its positive-excess
probability remains stable at 0.94.

\subsection{Summary-artifact replay}
\label{app:artifact-replay}

We replayed the ten readers from the original MuSiQue panel that could still be
run. Each received raw evidence, the main summary, and three fresh Qwen3 8B
summaries under one contemporaneous protocol.
Across artifacts, retention is 53.6--62.3\% under EM and 54.0--76.4\% under
F1. The main summary retains the most. Excess reversals are 31.6--52.2\pp{}
under EM and 14.8--40.0\pp{} under F1; all eight split intervals exclude zero.
The artifact draw therefore does not explain the excess-flip pattern.

The original fifteen-reader endpoint ratios are smaller because their raw
endpoint pairs are unavailable under the replay protocol. Claude 3.5 Haiku,
GLM-4 32B, Grok 4.1 Fast, and OLMo 3.1 32B are delisted, and their April runs
recorded no serving provider. DeepSeek R1 Distill 70B has one remaining
endpoint, but it ignores the reasoning-off request and returns no answer within
the fixed token budget. Recomputing the April main-summary scores on the
ten-reader subset gives 58.7\% EM and 52.3\% F1 retention. This provenance gap
is the same reason we require provider records for new replays.

\begin{table}[h]
\centering
\tablefont
\setlength{\tabcolsep}{4pt}
\begin{tabular}{@{}lrrrr@{}}
\toprule
Artifact & EM ret. & EM excess & F1 ret. & F1 excess \\
\midrule
Main & 62.3 & 31.6 & 76.4 & 14.8 \\
Draw 0 & 59.4 & 52.2 & 57.1 & 40.0 \\
Draw 1 & 60.9 & 50.0 & 54.0 & 36.0 \\
Draw 2 & 53.6 & 44.4 & 60.4 & 30.0 \\
\bottomrule
\end{tabular}
\caption{The MuSiQue attenuation and excess-flip patterns survive four summary
artifacts on the same ten-reader panel. Retention and excess flips are
percentages. Every excess-flip interval excludes zero. The artifacts are
correlated robustness draws within one benchmark.}
\label{tab:full-panel-artifact-replay}
\end{table}

An earlier replay generated the same three fresh caches over all 500 rows and
scored three Qwen3.5 readers. This replay provides provenance and matched-input
sensitivity evidence. Each reader also received a contemporaneous raw control.
Provider routing, prompts, temperature, seed, reasoning mode, and rows were
fixed. An abstaining summary remained compressed evidence; the runner never
replaced it with raw text.
The raw control is the released \texttt{naive\_top\_k} policy: it preserves
supplied order and passes the first eight paragraphs from the fixed candidate
pool of up to twenty paragraphs.

Table~\ref{tab:artifact-replay} selects endpoints from the raw scores. Qwen3.5
27B is strongest and 397B-A17B weakest, an order that differs from parameter
count. The raw F1 upgrade is 6.8\pp{}. All three regenerated artifacts attenuate
that upgrade while raising every reader by 19.5--22.8\pp{} F1. The 4.2\pp{}
raw EM upgrade is below the prespecified 5\pp{} ratio rule, so we leave EM retention
empty.

We then reran only the raw condition while passing every supplied paragraph to
the same three readers. Qwen3.5 9B and 27B become the raw endpoints. The raw EM
and F1 upgrades are both 10.6\pp{}. The summaries retain 8--11\% of the EM upgrade and
11--15\% of the F1 upgrade. This removes the candidate-pool-access asymmetry,
while length and token count still differ between raw text and summaries.

\begin{table}[h]
\centering
\tablefont
\setlength{\tabcolsep}{4pt}
\begin{adjustbox}{max width=\columnwidth}
\begin{tabular}{@{}lrrrrrr@{}}
\toprule
& \multicolumn{2}{c}{First-eight raw} & \multicolumn{4}{c}{All-paragraph raw} \\
\cmidrule(lr){2-3}\cmidrule(lr){4-7}
Draw & F1 comp. & F1 ret. & EM comp. & EM ret. & F1 comp. & F1 ret. \\
\midrule
0 & 4.3 & .64 & 1.0 & .09 & 1.2 & .11 \\
1 & 4.3 & .63 & 1.2 & .11 & 1.6 & .15 \\
2 & 3.5 & .52 & 0.8 & .08 & 1.6 & .15 \\
\bottomrule
\end{tabular}
\end{adjustbox}
\caption{One summary compressor attenuates the upgrade between the lowest- and
highest-scoring raw readers under both raw policies. Upgrades are percentage points. The first-eight F1 upgrade
is 6.8\pp{}; the all-paragraph EM and F1 upgrades are each 10.6\pp{}.
First-eight EM falls below the ratio threshold. Retention divides each visible
compressed upgrade by its raw upgrade.}
\label{tab:artifact-replay}
\end{table}

\paragraph{Artifact integrity.}
All three summaries are byte-identical on 385 of 500 rows (77.0\%). Pairwise
identity ranges from 80.8\% to 87.0\%. Every cache has 500 successful compiler
responses on the pinned endpoint and zero raw-evidence fallbacks. The original
replay contains 6,000 successful reader responses; the matched-input control
adds 1,500. Exact-prompt retries
repaired 90 initial empty reader responses and seven explicit upstream
rate-limit responses. Retries were limited to failures, and the artifact ledger
retains each retry.

\subsection{Prespecified predictive audit}
\label{app:predictive-audit}

Before evaluation, we froze a ranking-instability index that combines raw
reader spacing with compressor-class dispersion estimated without the held-out
cell. The spacing baseline has lower prediction error. Table~\ref{tab:predictive-audit}
collects the remaining diagnostics. We freeze this failed test without tuning.

\begin{table}[h]
\centering
\tablefont
\setlength{\tabcolsep}{4pt}
\begin{tabularx}{\columnwidth}{@{}lX@{}}
\toprule
Diagnostic & Result \\
\midrule
Leave-one-cell-out MAE & index .1005; spacing .0747 \\
Index-minus-spacing MAE & $+.0258\;[-.0099,.0615]$ \\
Disjoint-row median delta MAE & $+.0151\;[-.0025,.0384]$ across 5,000 splits \\
Benchmark-level errors improved & 2 of 4 \\
\bottomrule
\end{tabularx}
\caption{Raw-upgrade spacing outperforms the prespecified instability index. Brackets
are a benchmark-clustered interval for leave-one-cell-out error and a seeded
split range for the disjoint-row diagnostic.}
\label{tab:predictive-audit}
\end{table}

The four later LongMemEval readers test gain calibration. They do not test
cell-level ranking instability. Table~\ref{tab:temporal-prediction} reports absolute error
for three named gain baselines. DeepSeek-V4-Flash is also an explicit direction
miss: its recorded prediction is +3.0\pp{}, while its observed gain is
$-0.6$\pp{}.

\begin{table}[h]
\centering
\tablefont
\setlength{\tabcolsep}{3.5pt}
\begin{adjustbox}{max width=\columnwidth}
\begin{tabular}{@{}lrrrr@{}}
\toprule
Reader & Gain & Zero err. & Mean err. & Linear err. \\
\midrule
Qwen3.6 27B & +3.6 & 3.60 & 3.16 & .40 \\
DeepSeek-V4-Flash & $-0.6$ & .60 & 7.64 & 3.63 \\
GLM-5.1 & +1.6 & 1.60 & 4.97 & .63 \\
MiMo-v2.5 & +0.2 & .20 & 6.63 & 1.43 \\
\bottomrule
\end{tabular}
\end{adjustbox}
\caption{Temporal SIEVE-gain errors in percentage points. These readers are
holdouts evaluated after the original freeze.}
\label{tab:temporal-prediction}
\end{table}

\subsection{Deterministic reproduction}

The primary official-score analyses use seed 20260821 and 5,000 resamples; the
TriviaQA expansion uses seed 20260823 and the same resample count. The primary
score file contains 96,831 row-level metric records after the NQ exclusion. A
second complete scoring and analysis pass reproduces the score file, report,
and JSON hashes. The manuscript figures read the stored JSON artifact directly, except the
MuSiQue four-summary ranges in Figure~\ref{fig:scaling-collapse}, which are
copied from the replay report.

\subsection{Implementation validation of \texttt{ragscale}}
\label{app:ragscale-validation}

This is a software check. It adds no benchmark, reader, or compressor
evidence. Upgrade retention is the visible compressed upgrade of the reader
pair selected under raw evidence, divided by that pair's raw upgrade. Excess flips is the
compressed reversal rate minus the raw replay rate, taken as the median over
5,000 row splits.

The check rebuilds the \texttt{ragscale} input from the released scores, the
stored HotpotQA slice, and the cached RECOMP summaries. It
verifies the content hash of each source, uses the same random seeds as the
paper's analysis, and runs the package's replay audit without importing the
paper's analysis code. The package and the analysis were written by the same
authors from one specification. Agreement therefore shows that two
implementations compute the same function. This verifies implementation
consistency only.

\begin{table}[h]
\centering
\tablefont
\setlength{\tabcolsep}{4pt}
\begin{tabular}{@{}lrrrr@{}}
\toprule
Metric & Raw upgrade & Compressed upgrade & Retention & Excess flips (split range) \\
\midrule
EM & 31.8\pp{} & 7.8\pp{} & 0.245 & 16.7\pp{} [8.7, 24.4] \\
Token F1 & 34.7\pp{} & 8.5\pp{} & 0.245 & 16.7\pp{} [8.5, 24.1] \\
\bottomrule
\end{tabular}
\caption{\texttt{ragscale} reproduces the paper's HotpotQA RECOMP audit at
reported precision. Brackets give the 2.5 and 97.5 percentiles of excess
flips over 5,000 row splits. These split-sensitivity ranges do not estimate
population uncertainty. For EM the package also reproduces 19.1\% rescue and
35.7\% raw-correct damage.}
\label{tab:ragscale-validation}
\end{table}

Replacing one reader's artifact hash for one item makes the package refuse
the input before any statistic is computed. This guard covers only that
change. When a team supplies precomputed scores and hashes, the package can
only check that the hashes agree. When it runs the compressor itself, it
computes the hashes, so only that path enforces a fixed artifact.

The package also ships a key-only preset for illustration. It contributes no
evidence to the paper's claims. It runs the complete pipeline through a hosted API on 100
bundled rows of the same HotpotQA slice. A summary compressor produces one
fresh artifact per row, and two readers answer from it. This example confirms
that the package runs end to end against hosted models.

\section{Audit of Prior Compression Evaluations}
\label{app:prior-audit}

Across eight prior papers, primary cross-reader displays contain two to four
readers, with a median of three. None provides the row hashes or covariance
needed for our fixed-artifact EIV analysis
\citep{briefpro2026,exit2025,compact2024,casc2025,loocomp2026,hyco2_2025,
oscar2025,refiner2024}. Their conclusions are mixed: EXIT
reports larger strong-reader gains \citep{exit2025}, Refiner larger weak-reader
gains \citep{refiner2024}, and BRIEF-Pro large unanalyzed reader upgrades
\citep{briefpro2026}. The archived pooled correlations remain excluded. This
audit motivates the protocol while leaving the confirmatory benchmark count
unchanged.

\end{document}